\documentclass[cameraready]{Interspeech}

\usepackage{times}
\usepackage{latexsym}
\usepackage{inconsolata}
\usepackage{tikz}
\usepackage{colortbl,xcolor} 
\usepackage{arydshln} 
\usepackage{amsmath}
\usepackage{amssymb} 
\usepackage{multicol}
\usepackage{multirow}
\usepackage{cleveref}
\usepackage{booktabs}
\usepackage{siunitx}
\usepackage{graphicx}
\usepackage{fontawesome5}
\usepackage{subcaption}
\usepackage[most]{tcolorbox} 
\usepackage{tipa}
\usetikzlibrary{arrows.meta, positioning, shapes.geometric, calc}
\usepackage{fix-cm}
\usepackage{cite}
\newcommand{\lvsixty}{W2V2P-LV60\xspace}
\newcommand{\xlsrf}{W2V2P-XLSR53\xspace}

\newcommand{\zipactc}{ZIPA-CTC\xspace}
\newcommand{\zipactcns}{ZIPA-CTC-NS\xspace}
\newcommand{\powsm}{POWSM\xspace}
\newcommand{\powsmctc}{POWSM-CTC\xspace}

\newcommand{\huper}{HuPER-Recognizer\xspace}
\newcommand{\prism}{PRiSM\xspace}
\newcommand{\method}{PhoneticXEUS\xspace}
\newcommand{\ebformer}{E-Branchformer\xspace}
\newcommand{\wpad}{\textcolor{white!0}{0}}
\definecolor{colorint}{RGB}{214, 39, 40}
\definecolor{colorext}{RGB}{255, 176, 0}
\definecolor{colorother}{RGB}{44, 123, 182}

\usepackage{algorithm}
\usepackage{algpseudocode}
\usepackage{xcolor}
\usepackage{amsmath}

\title{An Empirical Recipe for Universal Phone Recognition}

\author[affiliation={1},orcid=0009-0000-4927-4561]{Shikhar}{Bharadwaj}
\author[affiliation={1},orcid=0009-0007-8221-4553]{Chin-Jou}{Li}
\author[affiliation={2},orcid=0000-0001-5254-1093]{Kwanghee}{Choi}
\author[affiliation={2},orcid=0009-0003-4452-4084]{Eunjung}{Yeo}
\author[affiliation={1},orcid=0000-0002-3251-3084]{William}{Chen}
\author[affiliation={1},orcid=0000-0002-5970-8631]{\\Shinji}{Watanabe}
\author[affiliation={1},orcid=0000-0002-3927-6851]{David R.}{Mortensen}

\address{
    $^1$ Carnegie Mellon University, USA \quad
    $^2$ The University of Texas at Austin, USA
}

\email{\{sbharad2,dmortens\}@andrew.cmu.edu}


\usepackage{comment}

\begin{document}

\maketitle

\begin{abstract}
Phone recognition (PR) is a key enabler of multilingual and low-resource speech processing tasks, yet robust performance remains elusive.
Highly performant English-focused models do not generalize across languages, while multilingual models underutilize pretrained representations.
It also remains unclear how data scale, architecture, and training objective contribute to multilingual PR.
We present PhoneticXEUS---trained on large-scale multilingual data and achieving state-of-the-art performance on both multilingual (17.7\% PFER) and accented English speech (10.6\% PFER).
Through controlled ablations with evaluations across 100+ languages under a unified scheme, we empirically establish our training recipe and quantify the impact of SSL representations, data scale, and loss objectives.
In addition, we analyze error patterns across language families, accented speech, and articulatory features.
All data and code are released openly.\footnote{\url{https://github.com/changelinglab/PhoneticXeus}.}
\end{abstract}

\section{Introduction}

\begin{figure}[t]
  \centering
  \includegraphics[width=1\linewidth]{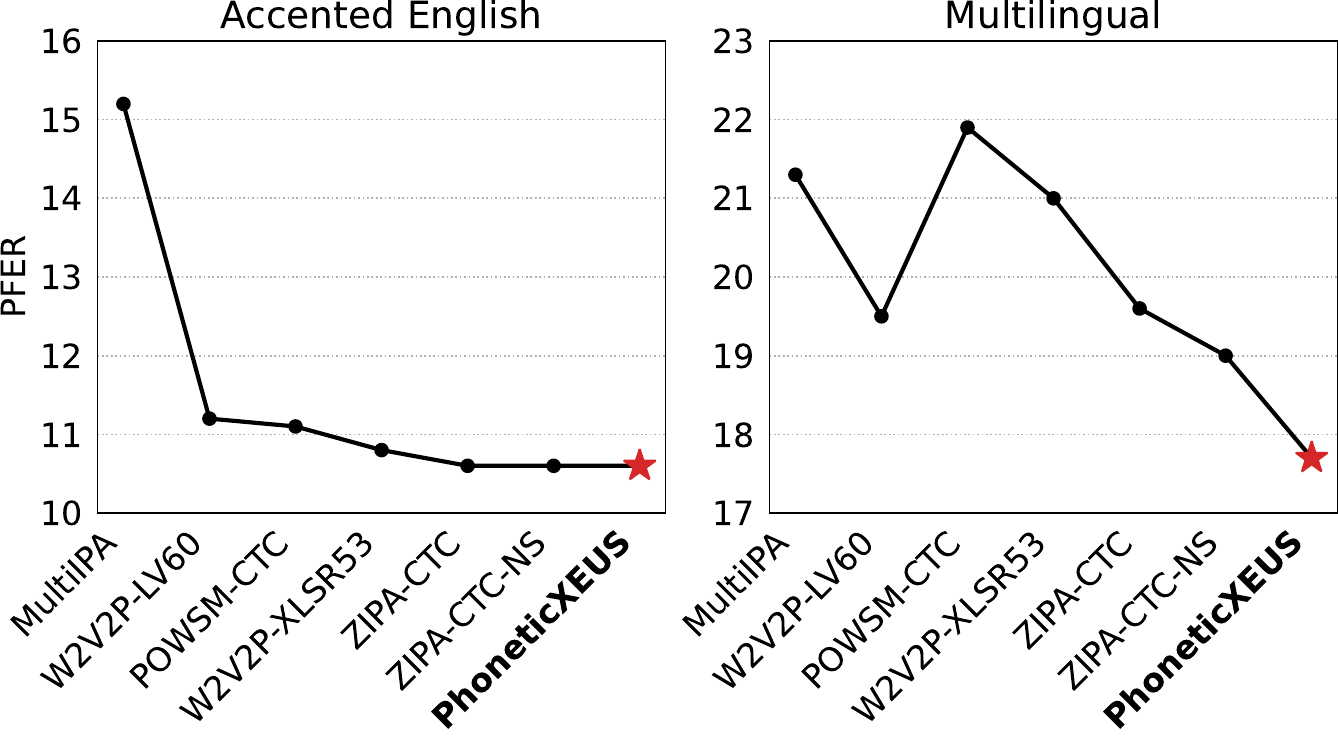}
  \caption{\method achieves SOTA performance on both accented English and multilingual speech. Details in \autoref{tab:phonebench:int-results}.}
  \label{fig:perfgap}
\end{figure}

Phone Recognition (PR) enables important multilingual speech processing technologies, especially for zero text-resource languages \cite{dunbar2019zero, dunbar2021zero, dunbar2022self} via cross-lingual transfer \cite{zelasko2020sounds, zhu2021multilingual, yan2021differentiable}.
PR is also heavily employed in atypical speech assessment \cite{lee2024learner, shriberg2025clinical}, computer-assisted language learning \cite{angkititrakul2006phone-based-accent, Franco2010EduSpeakAS, el2023automatic} and linguistic fieldwork \cite{chelliah2010handbook, ucla2009, mortensen2021tusom2021}.
For English PR, several systems have been developed since the 1950s \cite{dudley1958automatic, bhagath2004acoustic}, with recent systems \cite{huper, koellabs} pushing performance by leveraging larger and better-quality datasets.
However, their generalization to multilingual settings is limited (\Cref{fig:perfgap}).
On the other hand, recent progress has led to a proliferation of multilingual PR models and benchmarks \cite{zipa, powsm, prism}, yet these models still leave performance potential on the table by not fully leveraging self-supervised learning (SSL) based representations and by only exploring a narrow selection of training objectives.
Stated succinctly, \textbf{while English PR models do not generalize well to diverse languages, recent multilingual PR research has scarcely explored the space of possible training methods.}
To address this gap, we conduct extensive empirical experiments and establish a training recipe for a PR system that attains state-of-the-art performance in both unseen multilingual settings and on accented English.

\begin{table*}[tb!]
\caption{\method achieves SOTA performance on the \prism benchmark \cite{prism} across both accented English and multilingual settings. This table reports PFER ($\downarrow$) scores. \textcolor{gray}{Gray} denotes dataset included during training. $^{*}$ denotes uncertainty about training data.}
\label{tab:phonebench:int-results}
\centering
\resizebox{0.85\textwidth}{!}{%
\begin{tabular}{ll ccc >{\columncolor{gray!10}}c ccc >{\columncolor{gray!10}}c}
\toprule
 & & \multicolumn{4}{c}{\textbf{Accented English Datasets}} &
 \multicolumn{4}{c}{\textbf{Multilingual Datasets}} \\
\cmidrule(lr){3-6} \cmidrule(lr){7-10}

\textbf{Model} &  &
{\texttt{PR-tmt} } & {\texttt{PR-arc} } & {\texttt{PR-saa} }  & Avg. & {\texttt{PR-drc} } & {\texttt{PR-vox} } & {\texttt{PR-tsm} } & Avg. \\
\midrule
\textit{Large Audio Language Models} \\
Gemini 2.5 Flash$^{*}$ \cite{comanici2025gemini}   & & 15.2 & 12.7 & 13.2 & 13.7 & 105.3\wpad & 19.7 & 36.3  & 53.8 \\
Qwen3-Omni-Instruct$^{*}$ \cite{xu2025qwen3omnitechnicalreport}    & & 15.1 & 11.9 & \wpad9.1 & 12.0 & 150.2\wpad & 49.0 & 117.1\wpad  & 105.4\textcolor{gray!10}{0} \\
\midrule
\textit{English-centric PR Models} \\
KoelLabs-XLSR \cite{koellabs}     & & \wpad\textcolor{gray}{9.6} & \wpad\textcolor{gray}{7.8} & \wpad\textbf{7.8} & \textcolor{gray!10}{0}\textcolor{gray}{8.4} & 22.9 & 18.0 & 24.7 & 21.9 \\
\huper \cite{huper}        & & \wpad\textcolor{gray}{8.3} & 12.5 &  11.6 & \textcolor{gray}{{10.8}} & 32.0 &  26.5 & 29.8 & 28.2 \\
\midrule
\textit{Multilingual PR Models} \\
\lvsixty \cite{xu22b_interspeech}      & & \underline{13.2} & 10.9 & \wpad9.4 & 11.2 & 17.8 & 15.7 & 24.9 & 19.5 \\
\xlsrf \cite{xu22b_interspeech}        & & 13.5 & \wpad\underline{9.9} & \wpad9.0  & \underline{10.8} & 17.3 & \textbf{13.9} & 31.9 & 21.0 \\
MultiIPA  \cite{taguchi23_interspeech}      & & 16.3 & 15.5 & 13.8 & 15.2 & 18.3 & 15.2 & 30.5 & 21.3 \\
\zipactc \cite{zipa}      & & \textbf{13.1} & \textbf{\wpad9.7} & \wpad9.0 & \textbf{10.6} & 18.0 & 17.0 & 23.7  & 19.6 \\
\zipactcns \cite{zipa}    & & \textbf{13.1} & \textbf{\wpad9.7} & \wpad8.9 & \textbf{10.6} & \textbf{16.8} & 17.1 & 23.1  & 19.0 \\
\powsm  \cite{powsm}       & & 13.7 & 11.3 & 27.6 & 17.5 & \underline{17.1} & 17.1 & \underline{22.0} & \underline{18.7} \\
\powsmctc \cite{prism}     & & \textbf{13.1} & 10.3 & 10.0 & 11.1 & 18.1 &  15.3 & 32.2 & 21.9 \\
\rowcolor{colorint!10}\method (Ours)  & & 13.3 & \textcolor{colorint!10}{0}\underline{9.9} & \textcolor{colorint!10}{0}\underline{8.5} & \textbf{10.6} & \textbf{16.8} & \underline{14.4} & \textbf{21.9} & \textbf{17.7} \\

\bottomrule
\end{tabular}
}
\end{table*}

While prior approaches to multilingual PR train on a limited number of languages \cite{li2020towards, li2020universal, glocker23_interspeech, taguchi23_interspeech} or treated PR as a tool to explore unsupervised ASR (Wav2Vec2Phoneme) \cite{xu22b_interspeech}, recent methods \cite{zipa, powsm} have shown that scaling to diverse datasets \cite{zhu-etal-2024-taste} automatically generated via Grapheme-to-Phoneme (G2P) can significantly improve multilingual performance, allowing applications to real-world use cases \cite{yeo2026multilingual}.
However, more recent approaches do not utilize the capabilities of pretrained representations \cite{chen2024towards, pasad2021layer}.

Furthermore, the architecture and training objectives of prior systems have been chosen based their specific goals: 
ZIPA \cite{zipa} uses Zipformer \cite{yao2023zipformer} with CR-CTC \cite{crctc} for training efficiency, and \powsm \cite{powsm} uses CTC-Attention joint training in an autoregressive setup \cite{kim2017joint} for multi-task learning.
{Consequently, it remains unclear to what extent data scale, model architecture, and training objective contribute to multilingual PR performance.}
To cover these gaps, we provide controlled ablations of different components under \prism's \cite{prism} systematic evaluation scheme on human annotated datasets across more than 100 languages.
From these explorations, we build \method, a PR model with SOTA performance and robust multilingual capabilities.

Our experiments and analyses isolate the contribution of the various components of our recipe to performance on both seen and unseen languages. Further analyses explore the robustness of \method to cross-lingual transfer and accent variation as well as its effectiveness in recognizing articulatory features \cite{panphon}.
Our contributions are as follows:
\begin{itemize}
    \item We present \method, a state-of-the-art PR model, which excels at multilingual PR without degraded performance on accented English. To foster reproducibility, all the data and code for \method are open source.
    \item We provide detailed ablations to show the impact of each component of our training recipe toward final performance.
    \item We perform analyses to quantify the impact of SSL on cross-lingual transfer, report error patterns, and profile performance across diverse accents and articulatory features.
\end{itemize}

\section{Experiments}
We organize our experiments around three research questions: 
\begin{itemize}
    \item \textbf{RQ1}: Which CTC training objective best supports cross-lingual generalization in PR?
    \item \textbf{RQ2}: Do SSL representations pretrained on massively multilingual speech improve PR over training from scratch? 
    \item \textbf{RQ3}: How does the scale of multilingual training-data affect English versus multilingual performance?
\end{itemize}  
Combining the findings, we build \method, a PR model based on XEUS \cite{chen2024towards}---a large-scale multilingual pretrained SSL model---finetuned with Self-Conditioned CTC \cite{nozaki2021relaxing} on IPAPack++ \cite{zipa}, a multilingual phonemic dataset containing 17k hours of speech.
We evaluate the model with PRiSM \cite{prism}, a benchmark for PR systems, and report the results in Table~\ref{tab:phonebench:int-results}.
\method reached SOTA in accented English and greatly improved performance in multilingual datasets. 
All results are reported in PFER ($\downarrow$) unless otherwise noted.

\subsection{RQ1: On CTC and its variants}
\label{sec:rq1}

Encoder-CTC structures are shown to be more stable for PR \cite{prism}. 
Although several variants of the CTC loss have been proposed in the ASR literature \cite{graves2006connectionist,hierarchicalctc,lee2021intermediate, tjandra2020deja,nozaki2021relaxing, chen2023improving,kim2017joint}, they have been largely unexplored for phone recognition.
We evaluate five CTC-based loss functions, which we formally describe below.

\noindent\textbf{Notation.}
Let $\mathbf{x} \in \mathbb{R}^T$ be the input raw waveform of length $T$.
A PR model maps $\mathbf{x}$ to a reference phone sequence $\mathbf{y} = (y_1, \ldots, y_N)$ of length $N$, where each phone $y_n \in \mathcal{V}$ and $\mathcal{V}$ denotes the set of IPA symbols.

An encoder $f_\theta$ with learnable parameters $\theta$ and $M$ layers produces hidden representations
\[
\mathbf{H}^m = f_\theta^m(\mathbf{x}) = (\mathbf{h}^m_1, \ldots, \mathbf{h}^m_L) \in \mathbb{R}^{L \times D},
\]
for each layer index $m \in [1, M]$, where $L$ is the sequence length of the representations and $D$ is the hidden dimension.

\noindent\textbf{Vanilla CTC \cite{graves2006connectionist}.}
On top of $f_\theta$, we attach a learnable linear projection
$\mathbf{W}^M \in \mathbb{R}^{D \times (|\mathcal{V}|+1)}$
with bias
$\mathbf{b}^M \in \mathbb{R}^{|\mathcal{V}|+1}$,
followed by softmax function, to produce frame-level posteriors:
\begin{equation}
  \texttt{softmax}(\mathbf{W}^M \mathbf{H}^M + \mathbf{b}^M)
  = (\mathbf{p}^M_1, \ldots, \mathbf{p}^M_L)
  \in \mathbb{R}^{L \times (|\mathcal{V}|+1)}.
  \label{eq:posterior}
\end{equation}
We denote $\epsilon$ as the blank symbol, and the posterior probability of a symbol $v \in \mathcal{V} \cup \{\epsilon\}$ at frame $l$ as $\mathbf{p}^M_l[v]$.

The CTC loss marginalizes over all valid alignments
$\pi \in \Pi(\mathbf{y})$, where each alignment
$\pi = (\pi_1,\ldots,\pi_L)$ is a sequence with
$\pi_l \in \mathcal{V} \cup \{\epsilon\}$ that collapses to
$\mathbf{y}$ after removing blank symbols and merging consecutive duplicates.
The loss is defined as:
\begin{equation}
  \mathcal{L}^M_{\text{CTC}}(\mathbf{y} \mid \mathbf{x})
  = -\log \sum_{\pi \in \Pi(\mathbf{y})}
    \prod_{l=1}^{L} \mathbf{p}^M_l[\pi_l].
  \label{eq:ctc}
\end{equation}

\noindent\textbf{Intermediate CTC (InterCTC) \cite{hierarchicalctc, lee2021intermediate, tjandra2020deja}.}
Auxiliary CTC losses are applied to a subset of intermediate encoder layers
$\mathcal{S} \subset \{1, \ldots, M{-}1\}$.
For each layer $s \in \mathcal{S}$, a CTC loss is computed using the corresponding hidden representations $\mathbf{H}^s$.
The overall training objective is defined as
\begin{equation}
  \mathcal{L}_{\text{inter}}(\mathbf{y} \mid \mathbf{x})
  = \mathcal{L}_{\text{CTC}}^M(\mathbf{y} \mid \mathbf{x})
    + \lambda \cdot \frac{1}{|\mathcal{S}|}
      \sum_{s \in \mathcal{S}}
      \mathcal{L}_{\text{CTC}}^{s}(\mathbf{y} \mid \mathbf{x}),
  \label{eq:interctc}
\end{equation}
where $\mathcal{L}_{\text{CTC}}^{s}$ denotes the CTC loss computed from the representations of layer $s$, and $\lambda \in \mathbb{R}$ controls the contribution of the auxiliary losses.
The hyperparameter $\lambda \in \mathbb{R}$ is tuned on the validation set.
InterCTC regularizes intermediate representations and alleviates gradient vanishing in deeper layers.

\noindent\textbf{Self-Conditioned CTC (SelfCTC) \cite{nozaki2021relaxing, chen2023improving}.}
At each intermediate layer $s \in \mathcal{S}$, the phone posteriors $\mathbf{p}_l^{s}$ are projected through a learnable linear projection $\tilde{\mathbf{W}}^s \in \mathbb{R}^{D \times (|\mathcal{V}|+1)}$ and added to the hidden representations:
\begin{equation}
  \tilde{\mathbf{h}}_l^{s}
  = \mathbf{h}_l^{s} + \tilde{\mathbf{W}}^s \mathbf{p}_l^{s}.
  \label{eq:selfctc}
\end{equation}
The training objective is identical to \Cref{eq:interctc}.
This conditioning enables deeper layers to refine predictions by leveraging
phonetic context from earlier layer's predictions.

\noindent\textbf{Hierarchical CTC \cite{hierarchicalctc, higuchi2022hierarchical}.}
The loss is the same as in Self-Conditioned CTC, except that intermediate CTC layers operate on orthographic transcripts using a character vocabulary.

\noindent\textbf{Joint CTC-Attention \cite{kim2017joint}}.
To the CTC loss we add a cross-entropy loss $\mathcal{L}_{\text{CE}}$ from an autoregressive transformer decoder:
\begin{equation}
\mathcal{L}_{\text{CE}}(\mathbf{y}|\mathbf{x})
= - \sum_{n=1}^{N}
\log P(y_n | y_{<n} ;\mathbf{H}^{M}),
\end{equation}
where $P$ is obtained by applying a learnable linear projection and softmax to the decoder output representations.
The overall training objective combines CTC with CE:
\begin{equation}
  \lambda\mathcal{L}_{\text{CTC}}^{M}(\mathbf{y}|\mathbf{x})
    + (1-\lambda)\,\mathcal{L}_{\text{CE}}(\mathbf{y}|\mathbf{x}),
  \label{eq:jointctc}
\end{equation}
where $\lambda \in [0,1]$ balances the two objectives.
The CTC component provides alignment supervision while the attention decoder may further capture phonotactics.


\begin{table}[]
\centering
\caption{Ablation Study: CTC Loss Variations, finetuning XEUS (unless \colorbox{gray!10}{specified}) on IPAPack++. Our training recipe is \colorbox{colorint!10}{C3}.}
\label{tab:ablations-ctc}
\resizebox{0.95\columnwidth}{!}{%

\begin{tabular}{llcc}
\hline
& \textbf{CTC Loss} &  \textbf{English} & \textbf{Multiling.} \\
\hline
C1 & Vanilla CTC \cite{graves2006connectionist} & 10.5 & 18.8 \\
C2 & InterCTC \cite{lee2021intermediate}  & 10.5 & 18.5 \\
\rowcolor{colorint!10} C3 & SelfCTC \cite{nozaki2021relaxing} & 10.6 & 17.7 \\
C4 & \cellcolor{gray!10}\hspace{1em} on MMS (1B) & \cellcolor{gray!10} 10.7 & \cellcolor{gray!10} 19.7 \\
C5 & \cellcolor{gray!10}\hspace{1em} on \ebformer (580M) & \cellcolor{gray!10} 12.6 & \cellcolor{gray!10} 23.1 \\
C6 & Hierarchical CTC \cite{hierarchicalctc} & 10.5 & 18.9 \\
C7 & CTC-Attention \cite{kim2017joint} & 13.2 & 18.9 \\
C8 & \hspace{1em} inference w/o decoder & 10.8 & 21.1 \\
\hline
\end{tabular}

}%

\end{table}

\begin{table}[]
\centering
\caption{Ablation Study: SSL Representations and Model Architectures, using a subset of IPAPack++ with vanilla CTC loss.}
\label{tab:ablations-model}
\resizebox{0.95\columnwidth}{!}{
\begin{tabular}{llccc}
\hline
& \textbf{Backbone} & \textbf{Param.} & \textbf{English} & \textbf{Multiling.} \\
\hline
S1 & \ebformer \cite{kim2023branchformer} & 380M &  12.3 & 23.2 \\
S2 &\ebformer \cite{kim2023branchformer} & 580M & 12.6 & 23.1 \\
S3 &MMS \cite{pratap2023mms}       & 300M & 11.0 & 18.3 \\
S4 &MMS \cite{pratap2023mms}  & 1B   & 10.8 & 18.3 \\
S5 &XEUS \cite{chen2024towards}   & 580M   & 10.6 & 19.6 \\
\hline
\end{tabular}
}
\end{table}

\noindent\textbf{Results.}
In \Cref{tab:ablations-ctc}, all loss variants, except joint CTC-Attention (C7) produce similar English PFER (10.5 - 10.6), suggesting that most objectives do not affect performance on accented English.
However, multilingual PFER reveals meaningful differences: SelfCTC (C3) achieves 17.7, compared to 18.5 for InterCTC (C2) and 18.8 for vanilla CTC (C1).
The 1.1\% gain from self-conditioning suggests that feeding soft phonetic posteriors back into deeper encoder layers helps the model generalize across languages.
Orthographic supervision at intermediate layers using the Hierarchical CTC approach (C6) shows performance degradation on multilingual datasets.
For joint CTC-Attention, when we run decoding with only the encoder (C8) we see that performance on English is restored, but that on multilingual datasets decrease.
Based on these results, we adopt SelfCTC (C3) as the training objective in our recipe.

\begin{figure}[t]
  \centering
  \includegraphics[width=0.9\linewidth]{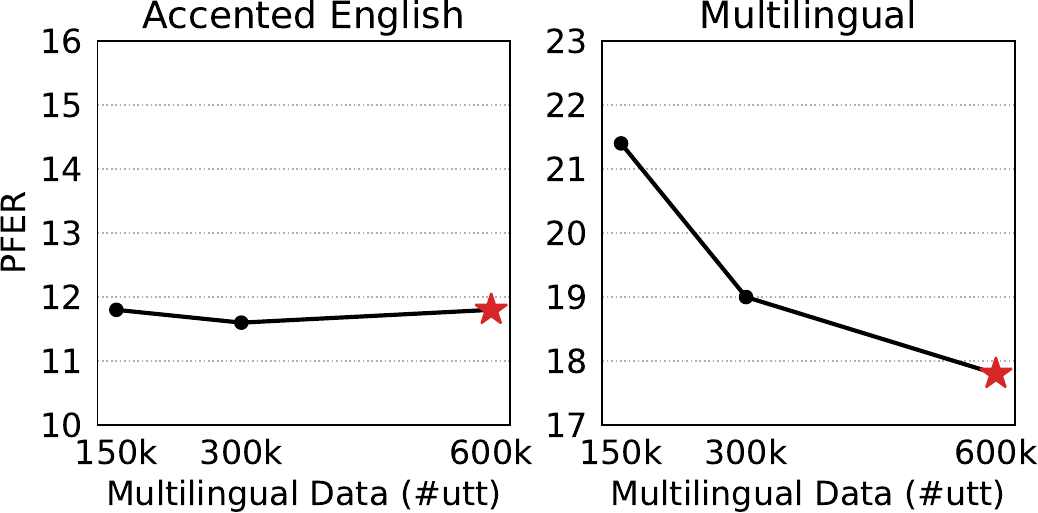}
  \caption{Increased language diversity in fine-tuning data benefits PR performance on multilingual datasets (\autoref{sec:rq3}).}
  \label{fig:langscale}
\end{figure}

\begin{figure*}[t]
  \centering
  \includegraphics[width=0.9\linewidth]{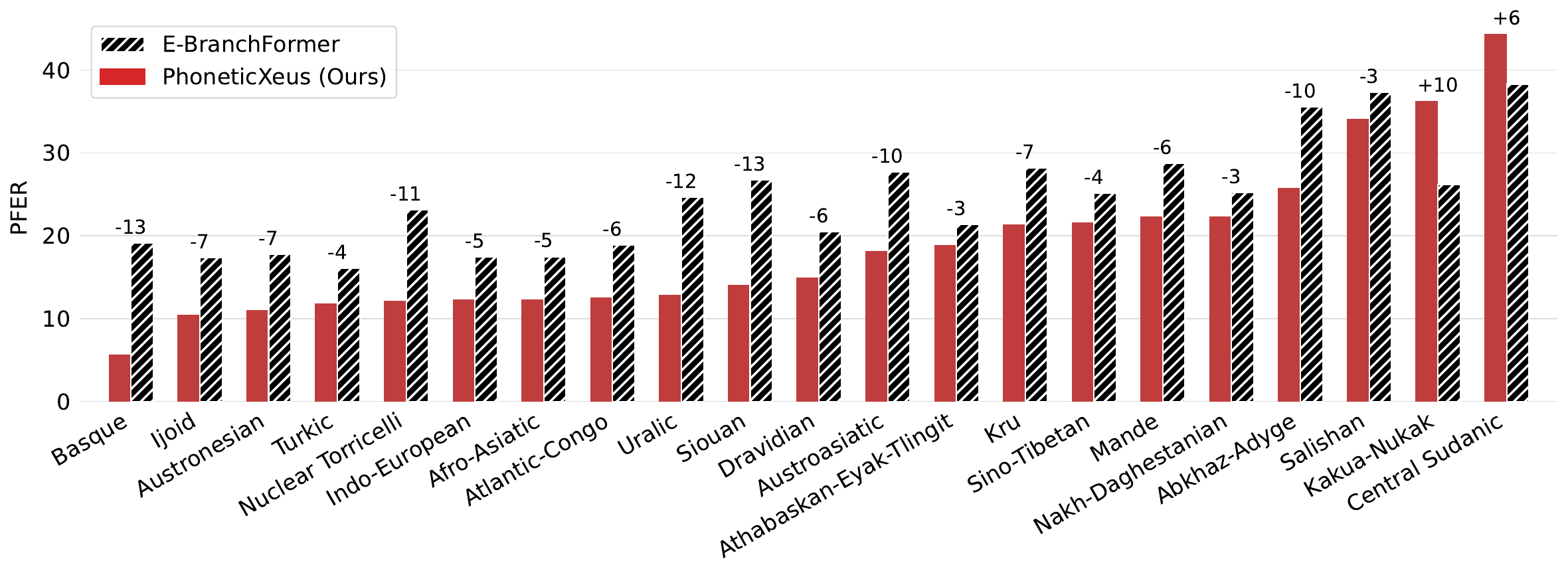}
  \vspace{-1em}
  \caption{Performance of \method across language families in VoxAngeles. SSL improves cross-lingual transfer (\autoref{sec:voxfamily}).}
  \label{fig:voxfamily}
\end{figure*}

\subsection{RQ2: On SSL Representations}
\label{sec:rq2}

Massively multilingual SSL encoders \cite{pratap2023mms,chen2024towards} learn representations that capture phonetic structure.
Recent PR models \cite{zipa, powsm} trained solely on G2P-transcribed speech might memorize standard pronunciation, ungrounded in acoustics.
We conjecture that SSL pretraining can ameliorate this issue, as they have been shown to encode phonological information \cite{choi2026self,choi2026orth}.
We train three configurations on a subset of IPAPack++:
1) \ebformer \cite{kim2023branchformer} (S1,S2) as baselines without pretraining.
2) MMS \cite{pratap2023mms} (S3, S4), pretrained on 1k languages.
3) XEUS \cite{chen2024towards} (S5), an \ebformer speech encoder trained with HuBERT-style \cite{hsu2021hubert} masked prediction on 4k languages.

In \autoref{tab:ablations-model}, MMS and XEUS both consistently show improvements over similar sized baselines (S1, S2).
We also experiment with MMS-1B on full IPAPack++ with SelfCTC (\autoref{tab:ablations-ctc} - C4) but we find that the out-of-domain multilingual performance of MMS deteriorates (S4 vs. C4).
However, XEUS achieves the best performance (C3) showing 2.0\% improvement on English and 5.4\% improvement on multilingual evaluations over the baseline trained from scratch (C5).
Hence, we adopt XEUS as the backbone in our recipe.
We further explore the effectiveness of SSL pretraining in \Cref{s:analyses}.

\subsection{RQ3: On Multilingual Training}
\label{sec:rq3}

Since G2P-generated labels are coarse and sometimes noisy, including phonemic contrasts from different languages might provide richer contrastive supervision across IPA inventory, and therefore average out bias from certain languages. 
Previous studies have shown that large-scale multilingual G2P data boosts performance \cite{zipa}, yet we do not know if gain plateaus and whether it affects high resource language performance.
We kept the number of English utterance the same (around 850k) and increase the number of utterance from other languages proportionally, from 150k to 300k to 600k. 
We trained \method with these data until convergence and evaluated the best checkpoint. 
Figure~\ref{fig:langscale} shows that including more multilingual data improves performance in multilingual datasets without damaging English performance, demonstrating the importance of dataset scale for multilingual PR.
Thus, we include all the multilingual data in our final recipe.

\begin{figure}[t]
  \centering
  \includegraphics[width=0.85\linewidth]{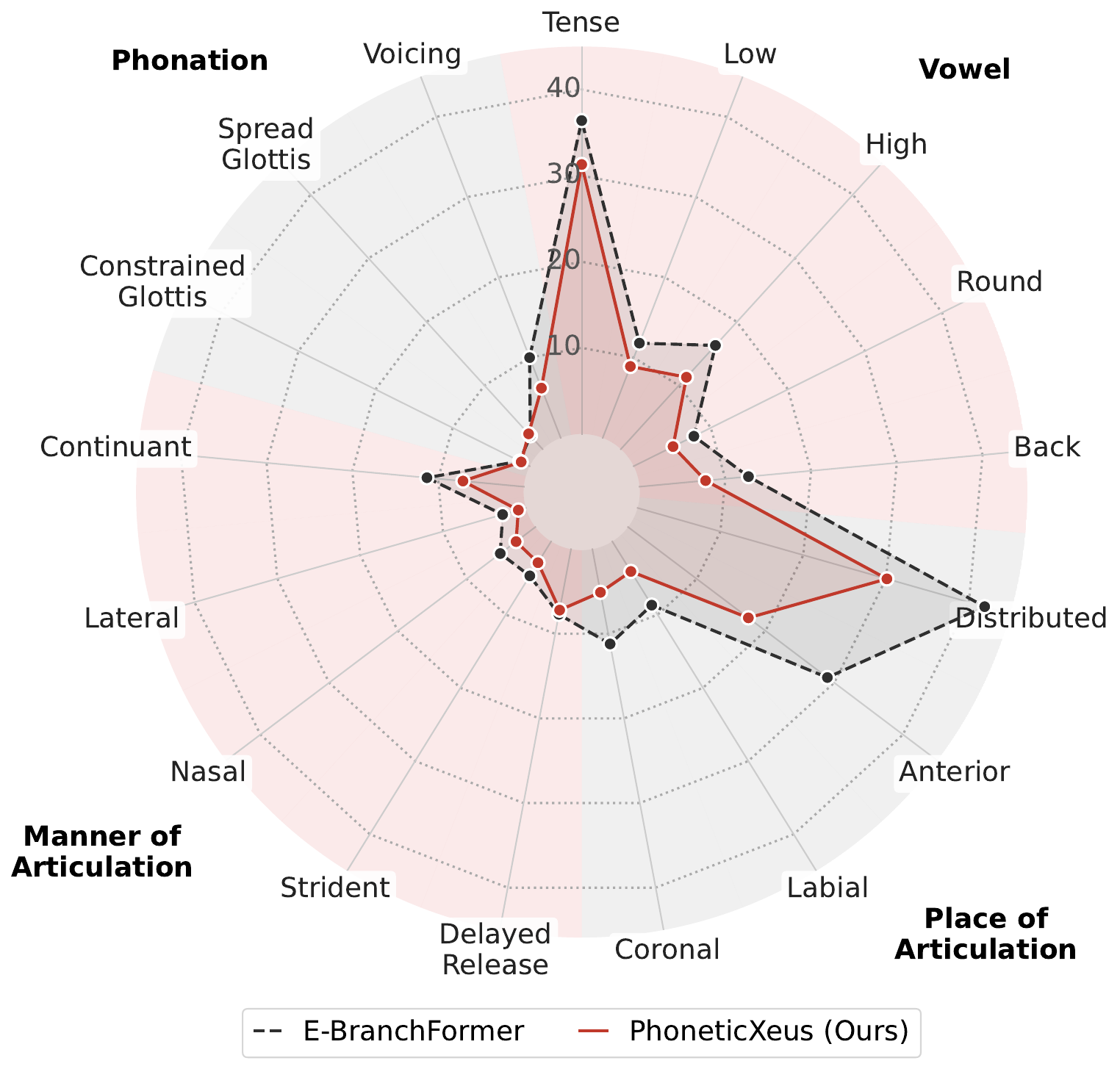}
  \vspace{-0.5em}
  \caption{Performance across articulatory features ($\downarrow$). Different features show different relative gains with SSL initialization. (\autoref{sec:artfeats}).}
  \label{fig:spider}
\end{figure}

\section{Analyses}\label{s:analyses}
Cross-lingual transfer is critical for PR in under-resourced languages.
We analyze the role of SSL in cross-lingual transfer via comparison with \ebformer (C5) which differs only in initialization to our recipe (C3).
Also, we aim to provide a deeper understanding of the model behavior and its limitations.

\subsection{Cross-lingual Performance and Failure Modes}
\label{sec:voxfamily}

\autoref{fig:voxfamily} shows that the use of SSL features improves performance in 19 of the 21 language families in \texttt{PR-vox} \cite{chodroff2024voxangeles} (covering 95 languages).
To estimate whether SSL representations allow the model to leverage training data more effectively, we conduct a rank correlation analysis between PFER scores per language and the language's coverage in our training data.
We estimate how well an unseen test language is represented in the PR training data by using similarity-weighted utterance counts.
To measure similarity between languages, we use phonological language vectors \cite{lang2vec}.
\method shows a consistent negative trend suggesting that better coverage is associated with lower error, reaching marginal significance ($\rho = -0.25$, $p = 0.096$), while \ebformer does not exhibit a meaningful relationship ($\rho = -0.09$, $p = 0.571$).
This suggests that SSL representations allow the model to better leverage phonological proximity to languages in paired training data for making predictions for unseen languages.

To better understand the limitations of our model, we next examine errors qualitatively in the lowest-performing languages.
We sample 103 utterances from the three lowest-performing languages: Lendu (\texttt{led}, 44\%), Wu Chinese (\texttt{wuu}, 40\%), and Kakua (\texttt{cbv}, 36\%).
For Lendu (23 utterances), utterances are monosyllabic words in a consonant-vowel pattern shorter than 1 second, and errors come from incorrect consonant (57\%), failed prediction (26\%), or mismatched annotation (17\%).
For Wu Chinese (40 utterances), the model either omitted glottal stops (40\%), suffered from ambiguous annotations (18\%), and did not predict (8\%).
For Kakua (40 utterances), we observed that all of the sampled utterances came from children or adult females and the errors stemmed from partial or approximate predictions.
Our initial analysis suggests that low performance could be attributed to a lack of robustness to acoustic shifts, in addition to noisy annotations.

\subsection{Performance Breakdown across Articulatory Features}
\label{sec:artfeats}
We additionally compare the performance of \method to \ebformer on articulatory features \cite{panphon} of \texttt{PR-vox}, by aligning the reference annotation with the model prediction, and report the proportion of errors for each feature (\autoref{fig:spider}).
We find that errors vary widely across attributes ranging from 5\% to 40\%, with SSL representations improving performance across all of them.
However, features characterized by temporal acoustic cues show the least benefit:
Tenseness (of vowels) and delayed release (of manner of articulation) show the lowest \textit{relative} decrease in error rate at 14\% and 6.5\% respectively, compared to over 50\% for features such as lateral and coronal.
Both features rely on on temporally distributed acoustic cues than on localized spectral patterns.
This analysis highlights weaknesses in current PR systems for capturing certain articulatory features, and future work can explore feature weighted losses to improve performance further.

\subsection{Performance on Accented English}
One concern with our training recipe is the use of G2P labels for training data, which are scalable but noisy:
G2P-generated labels reflect canonical dictionary pronunciations, which may not capture the phonetic variation present in accented or non-native speech \cite{zipa}.
To assess whether SSL representations confer robustness to accent variation despite this limitation, we compare \method and \ebformer on \texttt{PR-saa} \cite{gmuspeechaccentarchive}, a dataset with the same sentences spoken in multiple English accents.
\method improves over \ebformer across 187 of 192 accents (97\%), with an overall PFER reduction from 11.2\% to 8.8\% and gains as large as 6.3\% for Lao-accented English. 
This suggests that SSL pretraining provides robustness to accent variation that partially compensates for the canonical bias of G2P-based supervision. 
We leave a more principled approach to handling this mismatch to future work.
\section{Conclusion}

We present \method, a SOTA PR system along with systematic ablation study that disentangles the effects of training data, initialization, and training objective. 
Analyses reveal the role of SSL in cross-lingual transfer, and variance in performance across articulatory attributes, along with the need to develop better quality evaluation sets.

\section{Generative AI Use Disclosure}
Generative AI tools were used to improve the clarity and grammar of the manuscript and to assist with portions of the code.
All outputs were reviewed and validated by the authors.

\bibliographystyle{IEEEtran}
\bibliography{mybib}

\end{document}